\documentclass{article}
\usepackage{spconf,amsmath,graphicx}
\usepackage{graphicx}
\usepackage{amssymb,amsmath,bm}
\usepackage{textcomp}
\usepackage{multirow}
\usepackage{color}
\title{EESEN: End-to-End Speech Recognition using Deep RNN Models and WFST-based Decoding}
\name{Yajie Miao, Mohammad Gowayyed, Florian Metze}
\address{Language Technologies Institute, School of Computer Science, Carnegie Mellon University}
\begin{document}
%\ninept
%
\maketitle
\begin{abstract}
The performance of automatic speech recognition (ASR) has improved tremendously due to the application of deep neural networks (DNNs). Despite this progress, building a new ASR system remains a challenging task, requiring various resources, multiple training stages and significant expertise. This paper presents our Eesen framework which drastically simplifies the existing pipeline to build state-of-the-art ASR systems. Acoustic modeling in Eesen involves learning a single recurrent neural network (RNN) predicting context-independent targets (phonemes or characters). To remove the need for pre-generated frame labels, we adopt the connectionist temporal classification (CTC) objective function to infer the alignments between speech and label sequences. A distinctive feature of Eesen is a generalized decoding approach based on weighted finite-state transducers (WFSTs), which enables the efficient incorporation of lexicons and language models into CTC decoding. Experiments show that compared with the standard hybrid DNN systems, Eesen achieves comparable word error rates (WERs), while at the same time speeding up decoding significantly.  

\end{abstract}
\section{Introduction}
\label{sec:intro}

Automatic speech recognition (ASR) has traditionally leveraged the hidden Markov model/Gaussian mixture model (HMM/GMM) paradigm for acoustic modeling. HMMs act to normalize the temporal variability, whereas GMMs compute the emission probabilities of HMM states. In recent years, the performance of ASR has been improved dramatically by the introduction of deep neural networks (DNNs) as acoustic models \cite{dahl2012context, hinton2012deep, seide2011feature}. In the \textit{hybrid} HMM/DNN approach, DNNs are used to classify speech frames into clustered context-dependent (CD) states (i.e., senones). On a variety of ASR tasks, DNN models have shown significant gains over the GMM models. Despite these advances, building a state-of-the-art ASR system remains a complicated, expertise-intensive task. First, acoustic modeling typically requires various resources such as dictionaries and phonetic questions. Under certain conditions (e.g., in low-resource languages), these resources may be unavailable, which restricts or delays the deployment of ASR. Second, in the hybrid approach, training of DNNs still relies on GMM models to obtain (initial) frame-level labels. Building GMM models normally goes through multiple stages (e.g., CI phone, CD states, etc.), and every stage involves different feature processing techniques (e.g., LDA, fMLLR, etc.). Third, the development of ASR systems highly relies on ASR experts to determine the optimal configurations of a multitude of hyper-parameters, for instance, the number of senones and Gaussians in the GMM models. 

Previous work has made various attempts to reduce the complexity of ASR. In \cite{senior2014gmm, bacchiani2014asynchronous}, researchers propose to flat-start DNNs and thus get ride of GMM models. However, this GMM-free approach still requires iterative procedures such as generating forced alignments and decision trees. Meanwhile, another line of work \cite{graves2014towards, hannun2014deepspeech, hannun2014first, chorowski2014end, maas2015lexicon, bahdanau2015end, chan2015listen} has focused on end-to-end ASR, i.e., modeling the mapping between speech and labels (words, phonemes, etc.) directly without any intermediate components (e.g., GMMs). On this aspect, Graves \textit{et al.} \cite{graves2006connectionist} introduce the \textit{connectionist temporal classification} (CTC) objective function to infer speech-label alignments automatically. This CTC technique is further investigated in \cite{graves2014towards, hannun2014deepspeech, hannun2014first, sak2015learning} on large-scale acoustic modeling tasks. Although showing promising results, research on end-to-end ASR faces two major obstacles. First, it is challenging to incorporate lexicons and language models into decoding. When decoding CTC-trained models, past work \cite{graves2014towards, hannun2014first, maas2015lexicon} has successfully constrained search paths with lexicons. However, how to integrate \textit{word-level} language models efficiently still is an unanswered question \cite{maas2015lexicon}. Second, the community lacks a shared experimental platform for the purpose of benchmarking. End-to-end systems described in the literature differ not only in their model architectures but also in their decoding methods. For example, \cite{graves2014towards} and \cite{hannun2014first} adopt two distinct versions of beam search for decoding CTC models. These setup variations hamper rigorous comparisons not only across end-to-end systems, but also between the end-to-end and existing hybrid approaches.

In this paper, we resolve these issues by presenting and publicly releasing our Eesen framework. Acoustic modeling in Eesen is viewed as a sequence-to-sequence learning problem. We exploit deep recurrent neural networks (RNNs) \cite{graves2013speech, graves2013hybrid} as the acoustic models, and the Long Short-Term Memory (LSTM) units \cite{hochreiter1997long, sak2014long, sainath2015convolutional, miao2015on} as the RNN building blocks. Using the CTC objective function, Eesen simplifies acoustic modeling into learning a single RNN over pairs of speech and context-independent (CI) label sequences. A distinctive feature of Eesen is a generalized decoding method based on weighted finite-state transducers (WFSTs). In this method, individual components (CTC labels, lexicons and language models) are encoded into WFSTs, and then composed into a comprehensive search graph. The WFST representation provides a convenient way of handling the CTC \textit{blank} label and enabling beam search during decoding. Our experiments with the Wall Street Journal (WSJ) benchmark show that Eesen results in superior performance than the existing end-to-end ASR pipelines \cite{graves2014towards, hannun2014first}. The WERs of Eesen are on a par with strong hybrid HMM/DNN baselines. Moreover, the application of CI modeling targets allows Eesen to speed up decoding and reduce decoding memory usage. Eesen is released as an open-source project\footnote{https://github.com/yajiemiao/eesen}, and will undertake continuous expansion and optimization.

\section{The EESEN Framework: Model Training}
\label{sec:training}

Acoustic models in Eesen are deep bidirectional RNNs trained with the CTC objective function \cite{graves2006connectionist}. We describe the model structure in Section \ref{sec:training_1}, and restate key points of CTC training in Section \ref{sec:training_2}. Section \ref{sec:training_3} presents some practical considerations emerging from our GPU implementation. 

\subsection{Deep Bidirectional Recurrent Neural Networks} \label{sec:training_1}

Compared to the standard feedforward networks, RNNs have the advantage of learning complex temporal dynamics on sequences. Given an input sequence $\textbf{X} = (\textbf{x}_1, ..., \textbf{x}_T)$, a recurrent layer computes the \textit{forward} sequence of hidden states $\overrightarrow{\textbf{H}} = (\overrightarrow{\textbf{h}}_1, ..., \overrightarrow{\textbf{h}}_T)$  by iterating from $t=1$ to $T$:
\begin{equation}
  \overrightarrow{\textbf{h}}_{t} = \sigma(\overrightarrow{\textbf{W}}_{hx} \textbf{x}_{t} + \overrightarrow{\textbf{W}}_{hh} \overrightarrow{\textbf{h}}_{t-1} + \overrightarrow{\textbf{b}}_{h})
  \label{eq1}
\end{equation}
where $\overrightarrow{\textbf{W}}_{hx}$ is the input-to-hidden weight matrix, $\overrightarrow{\textbf{W}}_{hh}$ is the hidden-to-hidden weight matrix. In addition to the inputs $\textbf{x}_{t}$, the hidden activation $\textbf{h}_{t-1}$ from the previous time step are fed to influence the hidden outputs at the current time step. In a bidirectional RNN, an additional recurrent layer computes the \textit{backward} sequence of hidden outputs $\overleftarrow{\textbf{H}}$ from $t=T$ to $1$: 
\begin{equation}
  \overleftarrow{\textbf{h}}_{t} = \sigma(\overleftarrow{\textbf{W}}_{hx} \textbf{x}_{t} + \overleftarrow{\textbf{W}}_{hh} \overleftarrow{\textbf{h}}_{t-1} + \overleftarrow{\textbf{b}}_{h})
  \label{eq2}
\end{equation}
Our acoustic model is a deep architecture, in which we stack multiple bidirectional recurrent layers. At each frame $t$, the concatenation of the forward and backward hidden outputs $[\overrightarrow{\textbf{h}}_{t}, \overleftarrow{\textbf{h}}_{t}]$ from the current layer are treated as inputs into the next recurrent layer.

\begin{figure}
    \centering
    \includegraphics[scale=0.12]{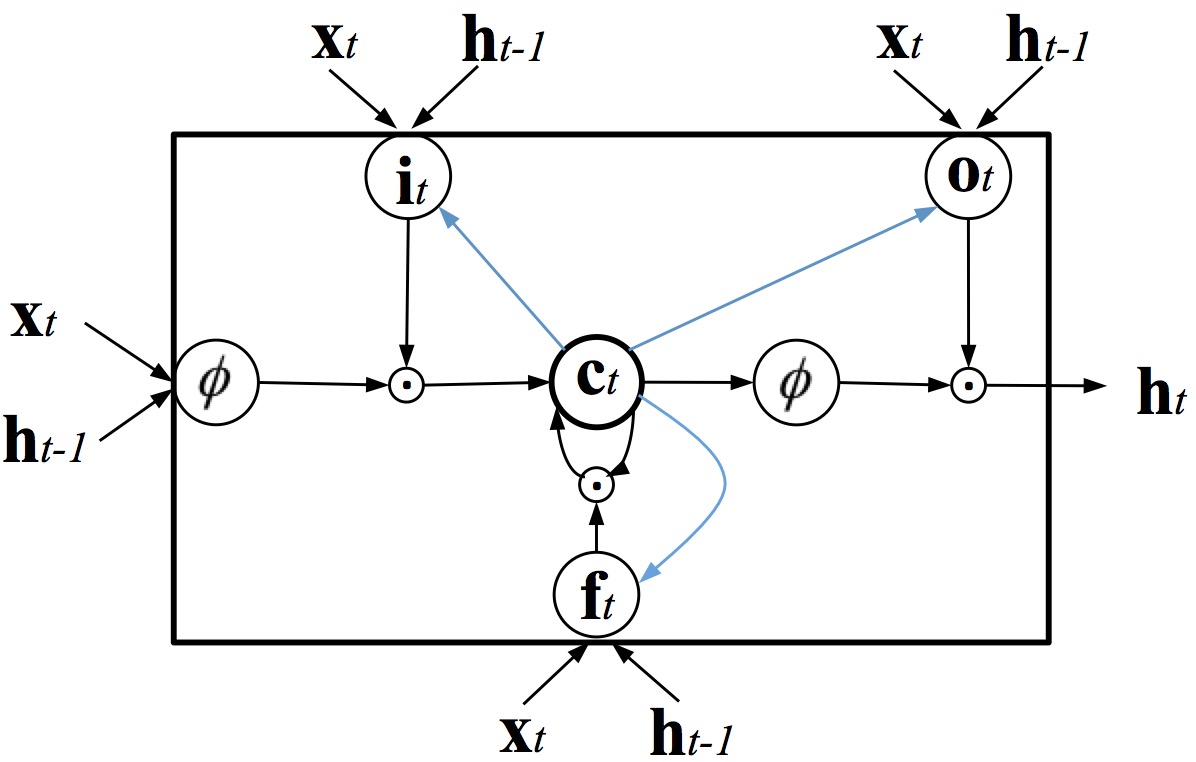}
    \caption{A memory block of LSTM. }
    \label{fig1}
    \vspace{-0.1in}
\end{figure}

Learning of RNNs can be done using back-propagation through time (BPTT). In practice, training RNNs to learn long-term temporal dependency can be difficult due to the vanishing gradients problem \cite{bengio1994learning}. To overcome this issue, we apply the LSTM units \cite{hochreiter1997long} as the building blocks of RNNs. LSTM contains memory cells with self-connections to store the temporal states of the network. Also, multiplicative gates are added to control the flow of information. Fig. \ref{fig1} depicts the structure of the LSTM units we use. The blue curves represent peephole connections \cite{gers2003learning} that link the memory cells to the gates to learn precise timing of the outputs.
The computation at the time step $t$ can be formally written as follows. We omit the $\rightarrow$ arrow for uncluttered formulation.   
\begin{subequations} \label{eq3}
  \begin{align}
  \textbf{i}_t = \sigma ( \textbf{W}_{ix} \textbf{x}_{t} + \textbf{W}_{ih} \textbf{h}_{t-1} + \textbf{W}_{ic} \textbf{c}_{t-1} + \textbf{b}_{i}) \label{eq3a}\\
  \textbf{f}_t = \sigma ( \textbf{W}_{fx} \textbf{x}_{t} + \textbf{W}_{fh} \textbf{h}_{t-1} + \textbf{W}_{fc} \textbf{c}_{t-1} + \textbf{b}_{f}) \label{eq3b}\\
  \textbf{c}_t = \textbf{f}_t \odot \textbf{c}_{t-1} + \textbf{i}_t \odot \phi( \textbf{W}_{cx} \textbf{x}_{t} + \textbf{W}_{ch} \textbf{h}_{t-1} + \textbf{b}_{c}) \label{eq3c} \\
  \textbf{o}_t = \sigma ( \textbf{W}_{ox} \textbf{x}_{t} + \textbf{W}_{oh} \textbf{h}_{t-1} + \textbf{W}_{oc} \textbf{c}_{t} + \textbf{b}_{o}) \label{eq3d} \\
  \textbf{h}_t = \textbf{o}_{t} \odot \phi(\textbf{c}_t) \label{eq3e}
  \end{align}
\end{subequations}
where $\textbf{i}_t$, $\textbf{o}_t$, $\textbf{f}_t$, $\textbf{c}_t$ are the activation of the input gates, output gates, forget gates and memory cells respectively. The $\textbf{W}_{.x}$ weight matrices connect the inputs with the units, whereas the $\textbf{W}_{.h}$ matrices connect the \textit{previous} hidden states with the units. The $\textbf{W}_{.c}$ terms are diagonal weight matrices for peephole connections. Also, $\sigma$ is the logistic sigmoid nonlinearity, and $\phi$ is the hyperbolic tangent nonlinearity. The computation of the \textit{backward} LSTM layer can be represented similarly. In this work, we use a purely LSTM-based architecture as the acoustic model. However, combing LSTMs with other network structures, e.g., time-delay \cite{waibel1989phoneme, hampshire1990novel} or convolutional neural networks \cite{sainath2014deep, sainath2015convolutional}, is straightforward to achieve.  

\subsection{Training with Connectionist Temporal Classification} \label{sec:training_2}

Unlike in the hybrid approach, the RNN model in our Eesen framework is not trained using frame-level labels with respect to the cross-entropy (CE) criterion. Instead, following \cite{graves2014towards, hannun2014first, maas2015lexicon}, we adopt the CTC objective \cite{graves2006connectionist} to automatically learn the alignments between speech frames and their label sequences (e.g., phonemes or characters). Assume that the label sequences in the training data contain $K$ unique labels. Normally $K$ is a relatively small number, e.g., around 45 for English when the labels are phonemes. An additional \textit{blank} label $\varnothing$, which means no labels being emitted, is added to the labels. For simplicity of formulation, we denote every label using its index in the label set. Given an utterance $\textbf{X} = (\textbf{x}_1, ..., \textbf{x}_T)$, its label sequence is denoted as $\textbf{z} = (z_1, ..., z_U)$. In our implementation, we always index the blank as 0. Therefore $\textbf{z}_u$ is an integer ranging from 1 to $K$. The length of $\textbf{z}$ is constrained to be no greater than the length of the utterance, i.e., $U \leq T$. CTC aims to maximize $\ln Pr(\textbf{z} | \textbf{X})$, the log-likelihood of the label sequence given the inputs, by optimizing the RNN model parameters.

The final layer of the RNN is a softmax layer which has $K+1$ nodes that correspond to the $K+1$ labels (including $\varnothing$). At each frame $t$, we get the output vector $\textbf{y}_{t}$ whose $k$-th element $y_{t}^{k}$ is the posterior probability of the label $k$. However, since the labels $\textbf{z}$ are not aligned to the frames, it is difficult to evaluate the likelihood of $\textbf{z}$ given the RNN outputs. To bridge the RNN outputs with label sequences, an intermediate representation, the \textit{CTC path}, is introduced in \cite{graves2006connectionist}. A CTC path $\textbf{p} = (p_1, ..., p_T)$ is a sequence of labels at the frame level. It differs from $\textbf{z}$ in that the CTC path allows occurrences of the blank label and repetitions of non-blank labels. The total probability of the CTC path is decomposed into the probability of the label $p_t$ at each frame:
\begin{equation}
  Pr(\textbf{p} | \textbf{X}) = \prod_{t=1}^{T} y_{t}^{p_t} 
  \label{eq4}
\end{equation}
The label sequence $\textbf{z}$ can then be mapped to its corresponding CTC paths. This is a one-to-multiple mapping because multiple CTC paths can correspond to the same label sequence. For example, both ``A A $\varnothing$ $\varnothing$ B C $\varnothing$'' and ``$\varnothing$ A A B $\varnothing$ C C'' are mapped to the label sequence ``A B C''. We denote the set of CTC paths for $\textbf{z}$ as $\Phi(\textbf{z})$. Then, the likelihood of $\textbf{z}$ can be evaluated as a sum of the probabilities of its CTC paths:
\begin{equation}
  Pr(\textbf{z} | \textbf{X}) = \sum_{p \in \Phi(\textbf{z})}  Pr(\textbf{p} | \textbf{X})
  \label{eq5}
\end{equation}
However, summing over all the CTC paths is computationally intractable. A solution is to represent the possible CTC paths compactly as a trellis. To allow blanks in CTC paths, we add ``0'' (the index of $\varnothing$) to the beginning and the end of $\textbf{z}$, and also insert ``0'' between every pair of the original labels in $\textbf{z}$. The resulting \textit{augmented label sequence}  $\textbf{l} = (l_1, ..., l_{2U+1})$ is leveraged in a forward-backward algorithm for efficient likelihood evaluation. Specifically, in a forward pass, the variable $\alpha_{t}^{u}$ represents the total probability of all CTC paths that end with label $l_u$ at frame $t$. As with the case of HMMs \cite{rabiner1989tutorial}, $\alpha_{t}^{u}$ can be recursively computed from $\alpha_{t-1}^{u}$ and $\alpha_{t-1}^{u-1}$. Similarly, a backward variable $\beta_{t}^{u}$ carries the total probability of all CTC paths that starts with label $l_u$ at $t$ and reaches the final frame $T$. The likelihood of the label sequence $\textbf{z}$ can then be computed as:
\begin{equation}
  Pr(\textbf{z} | \textbf{X}) = { \sum_{u=1}^{2U+1}  {  \alpha_{t}^{u} \beta_{t}^{u}} }
  \label{eq6}
\end{equation}
where $t$ can be any frame $ 1 \leq t \leq T$. The objective $\ln Pr(\textbf{z} | \textbf{X})$ now becomes differentiable with respect to the RNN outputs $\textbf{y}_{t}$. We define an operation on the augmented label sequence $\Upsilon(\textbf{l}, k) = \{ u | l_u = k \}$ that returns the elements of $\textbf{l}$ which have the value $k$. The derivative of the objective with respect to ${y}_{t}^{k}$ can be derived as:
\begin{equation}
 {\partial \ln Pr(\textbf{z} | \textbf{X})  \over \partial  y_{t}^{k} } = {1 \over Pr(\textbf{z} | \textbf{X})}  {1 \over y_{t}^{k} } \sum_{u \in \Upsilon(\textbf{l}, k)}  \alpha_{t}^{u} \beta_{t}^{u} 
  \label{eq7}
\end{equation}
These errors are back-propagated through the softmax layer and further into the RNN to update the model parameters.  

\subsection{GPU Implementation} \label{sec:training_3}

We implement the training of the RNN models on GPU devices. To fully exploit the capacity of GPUs, multiple utterances are processed at a time in parallel. This parallel processing speeds up model training by replacing  matrix-vector multiplication over single frames with matrix-matrix multiplication over multiple frames. Within a group of parallel utterances, we pad every utterance to the length of the longest utterance in the group. These padding frames are excluded from gradients computation and parameter updating. For further  acceleration, the training utterances are sorted by their lengths, from the shortest to the longest. The utterances in the same group then have approximately the same length, which minimizes the number of padding frames.  To ensure training stability, the gradients of RNN parameters are clipped to the range of [-50, 50].

CTC learning is also expensive because the forward and backward vectors ($ \boldsymbol{ \alpha }_{t}$ and $\boldsymbol{\beta}_{t}$) have to be computed sequentially, either from $t=1$ to $T$ or from $t=T$ to $1$. Like in RNNs, our implementation of CTC also processes multiple utterances at the same time. Moreover, at a specific frame $t$, the elements of $\boldsymbol{\alpha}_{t}$ (and $\boldsymbol{\beta}_{t}$) are independent and thus can be computed in parallel.  

\section{The EESEN Framework: Decoding}
\label{sec:decoding}

\subsection{Decoding with WFSTs} \label{sec:decoding_1}

Previous work has introduced a variety of methods \cite{graves2014towards, hannun2014first, maas2015lexicon} to decode CTC-trained models. These methods, however, either fail to integrate word-level language models \cite{maas2015lexicon} or achieve the integration under constrained conditions (e.g., n-best list rescoring in \cite{graves2014towards}). In this work, we propose a generalized decoding approach based on WFSTs \cite{mohri2002weighted, povey2011kaldi}. A WFST is a finite-state acceptor (FSA) in which each transition has an input symbol, an output symbol and a weight. A path through the WFST takes a sequence of input symbols and emits a sequence of output symbols. Our decoding method represents the CTC labels, lexicons and language models as separate WFSTs. Using highly-optimized FST libraries such as OpenFST \cite{allauzen2007openfst}, we can fuse the WFSTs efficiently into a single search graph. Building of the individual WFSTs is described as follows. Although exemplified in the scenario of English, the same procedures hold for other languages. 

\begin{figure}
    \centering
    \includegraphics[scale=0.14]{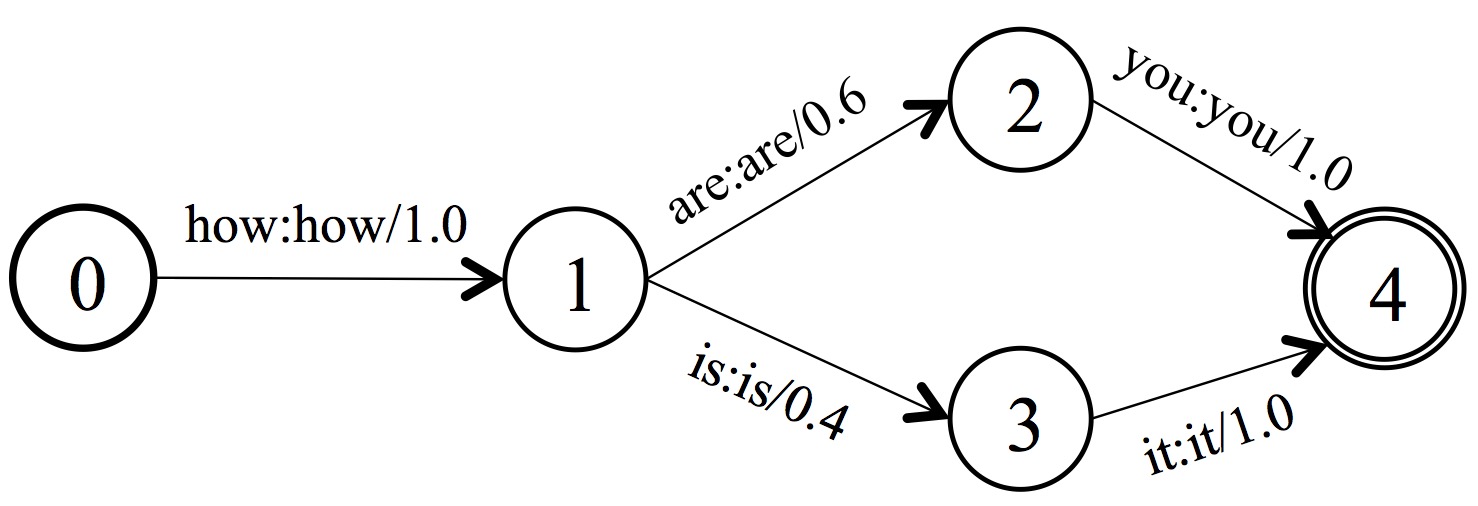}
    \caption{A toy example of the grammar (language model) WFST. The arc weights are the probability of emitting the next word when given the previous word. The node 0 is the start node, and the double-circled node is the end node. }
    \label{fig:gfst}
    \vspace{-0.1in}
\end{figure}

\textbf{Grammar}. A grammar WFST encodes the permissible word sequences in a language/domain. The WFST shown in Fig. \ref{fig:gfst} represents a toy language model which permits two sentences ``how are you'' and ``how is it''. The WFST symbols are the words, and the arc weights are the language model probabilities. With this WFST representation, CTC decoding in principle can leverage any language models that can be converted into WFSTs. Following conventions in the literature \cite{povey2011kaldi}, the language model WFST is denoted as $G$. 

\begin{figure}
    \centering
    \includegraphics[scale=0.13]{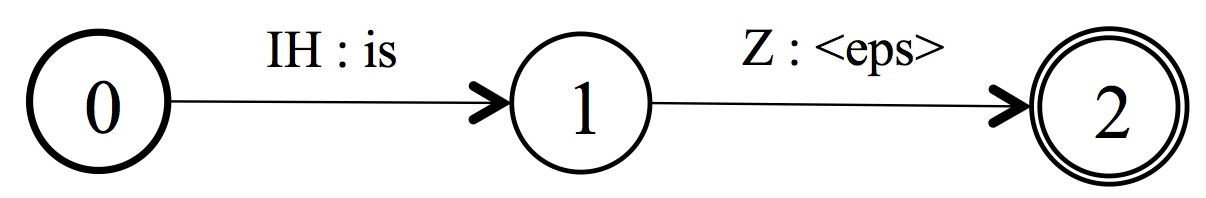}
    \caption{The WFST for the phoneme-lexicon entry ``is IH Z''. The ``$<$eps$>$'' symbol means no inputs are consumed or no outputs are emitted. }
    \label{fig:lpfst}
\end{figure}

\begin{figure}
    \centering
    \includegraphics[scale=0.13]{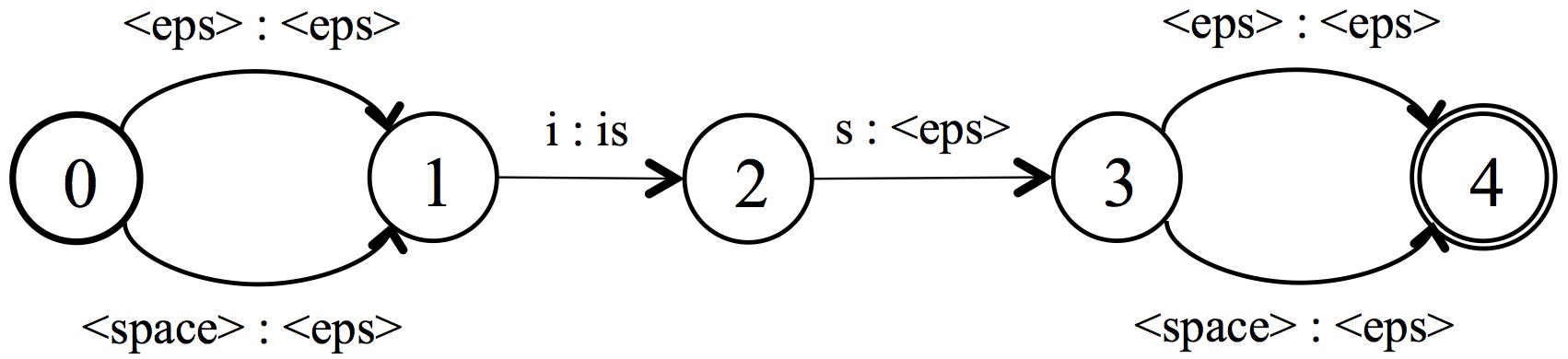}
    \caption{The WFST for the spelling of the word ``is''. We allow the word to optionally start and end with the space character ``$<$space$>$''. }
    \label{fig:lcfst}
    \vspace{-0.1in}
\end{figure}

\textbf{Lexicon}. A lexicon WFST encodes the mapping from sequences of lexicon units to words. Depending on what labels our RNN has modeled, there are two cases to consider. If the labels are phonemes, the lexicon is a standard dictionary as we normally have in the hybrid approach. When the labels are characters, the lexicon simply contains the spellings of the words. A key difference between these two cases is that the \textit{spelling lexicon} can be easily expanded to include any out-of-vocabulary (OOV) words. In contrast, expansion of the \textit{phoneme lexicon} is not so straightforward. It relies on some grapheme-to-phoneme rules/models, and is potentially subject to errors. The lexicon WFST is denoted as $L$.  Fig. \ref{fig:lpfst} and \ref{fig:lcfst} illustrate these two cases of building $L$.  

For the spelling lexicon, there is another complication to deal with. With characters as CTC labels, we usually insert an additional \textit{space} character between every pair of words, in order to model word delimiting in the original transcripts. During decoding, we allow the space character to optionally appear at the beginning and end of a word. This complication can be handled easily by the WFST shown in Fig. \ref{fig:lcfst}.

\begin{figure}
    \centering
    \includegraphics[scale=0.13]{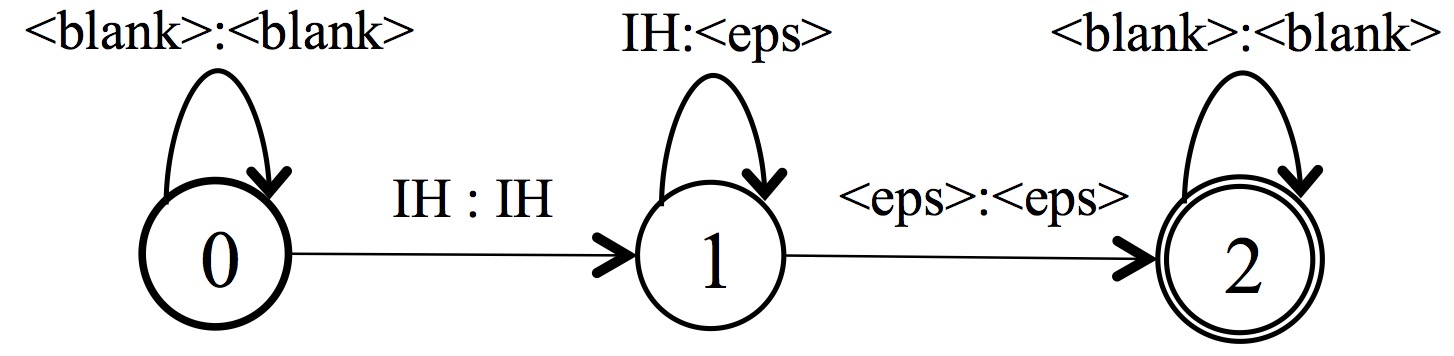}
    \caption{An example of the token WFST which depicts the phoneme ``IH''. We allow the occurrences of the blank label ``$<$blank$>$'' and the repetitions of the non-blank label ``IH''. }
    \label{fig:tfst}
\end{figure}

\textbf{Token}. The third WFST component maps a sequence of frame-level CTC labels to a single lexicon unit (phoneme or character). For a lexicon unit, its \textit{token WFST} is designed to subsume all of its possible label sequences at the frame level. Therefore, this WFST allows  occurrences of the blank label $\varnothing$, as well as repetitions of any non-blank lables. For example, after processing 5 frames, the RNN model may generate 3 possible label sequences  ``AAAAA'', ``$\varnothing$ $\varnothing$ A A $\varnothing$'', ``$\varnothing$ A A A $\varnothing$''. The token WFST maps all these 3 sequences into a singleton lexicon unit ``A''. Fig. \ref{fig:tfst} shows the WFST structure for the phoneme "IH". We denote the token WFST as $T$. 

\textbf{Search Graph}. After compiling the three individual WFSTs, we compose them into a comprehensive search graph. The lexicon and grammar WFSTs are firstly composed. Two special WFST operations,  \textit{determinization} and \textit{minimization}, are performed over the composition of them, in order to compress the search space and thus speed up decoding. The resulting WFST $LG$ is then composed with the token WFST, which finally generates the search graph. Overall the oder of the FST operations is:

\begin{equation}
  S = T \circ min( det (L \circ G) )
  \label{eq3}
\end{equation}
where $\circ$, $det$ and $min$ denote composition, determinization and minimization respectively. The search graph $S$ encodes the mapping from a sequence of CTC labels emitted on speech frames to a sequence of words. 

\subsection{Posterior Normalization}  \label{sec:decoding_2}

When decoding the hybrid DNN models, we need to scale the states posteriors from the DNNs using states priors. The priors are usually estimated from the forced alignments of the training data. During decoding of the CTC-trained models, we adopt a similar procedure. Specifically, we run the final RNN model over the training set for a propagation pass. Labels with the largest posteriors are picked as the frame-level alignments, from which priors of the labels are estimated. However, this method does not perform well in our experiments. Part of the reason is that the softmax-layer outputs from a CTC-trained model display a highly peaky distribution \cite{graves2006connectionist, sak2015learning}. That is, a majority of the frames have the blank as their labels. The activation of the non-blank labels only appears in a narrow region along the time axis. This causes the prior estimates to be dominated by the count of the blank. 

Alternatively, we propose to estimate more robust label priors from the label sequences in the training data. As mentioned in Section \ref{sec:training_2}, the label sequences actually used by CTC training are the augmented label sequences, which insert the blank at the beginning, at the end, and between every label pair  in the original label sequences. We compute the priors from the augmented label sequences (e.g., "$\varnothing$ IH $\varnothing$ Z $\varnothing$"), instead of the original ones (e.g., "IH Z"), through simple counting. In our experiments, this simple method gives better recognition accuracy than both the aforementioned frame-alignment method and also the proposal described in \cite{sak2015learning}. 

\section{Experiments}  \label{sec:exp}

\subsection{Experimental Setup} \label{sec:exp_1}
The experiments are conducted on the Wall Street Journal (WSJ) corpus that can be obtained from LDC under the catalog numbers LDC93S6B and LDC94S13B. Data preparation gives us 81 hours of transcribed speech, from which we select 95\% as the training set and the remaining 5\% for cross validation. As discussed in Section \ref{sec:training}, we apply deep RNNs as the acoustic models. Inputs of the RNNs are 40-dimensional filterbank features together with their first and second-order derivatives. The features are normalized via mean subtraction and variance normalization on the speaker basis.    

Initial values of the model parameters are randomly drawn from a uniform distribution with the range [$-$0.1, 0.1]. The model is trained with BPTT, in which the errors are back-propagated from CTC. Utterances in the training set are sorted by their lengths, and 10 utterances are processed in parallel at a time. The error rate of the \textit{hypothesized labels} is monitored to determine learning rates. The hypothesized labels are formed by firstly picking the frame-level labels (the label with the largest probability at every frame), and then removing blanks and label repetitions. The \textit{label error rate} (LER) can be obtained in the same manner as WER, i.e., computing the edit distance between the hypothesized labels and the reference. We adopt a decaying ``newbob'' learning rate schedule based on LERs. Specifically, the learning rate starts from 0.00004 and remains unchanged until the drop of LER on the validation set between two consecutive epochs falls below 0.5\%. Then the learning rate is decayed by a factor of 0.5 at each of the subsequent epochs. The whole learning process terminates when the LER fails to decrease by 0.1\% between two successive epochs. 

Our decoding follows the WFST-based approach in Section \ref{sec:decoding}. After posterior normalization, the acoustic model scores need to be scaled down. The scaling factor lies between 0.5 and 0.9, and its optimal value is decided empirically. We apply the WSJ standard pruned trigram language model in the ARPA format (which we will consistently refer to as \textit{standard}). To be consistent with previous work \cite{graves2014towards, hannun2014first}, we report our results on the \textit{eval92} set. Our experimental setup has been released together with Eesen, which enables the readers to reproduce our numbers easily.  

\subsection{Phoneme-based Systems} \label{sec:exp_2}

We explore the optimal RNN configurations on the phoneme-based systems. When phonemes are taken as CTC labels, we employ the CMU dictionary\footnote{http://www.speech.cs.cmu.edu/cgi-bin/cmudict} as the lexicon. Due to the lack of forced alignments, CTC training cannot handle multiple pronunciations for the same word. For every word, we only keep its first pronunciation in the lexicon and remove all the other alternatives. From the lexicon, we extract 72 labels including phonemes, noise marks and the blank. Our best-performing model has 4 bi-directional LSTM layers. At each layer, both the forward and the backward sub-layers contain 320 memory cells. Model training ends up to reach the LER (phone error rate in this setting) of 8.8\% on the validation set. On the eval92 testing set, the Eesen end-to-end system finally achieves the WER of 7.87\%, with both the lexicon and the language model used in decoding. When only the lexicon is used, our decoding behaves similarly as the beam search in \cite{graves2014towards}. In this case, the WER rises quickly to 26.92\%. This obvious degradation reveals the effectiveness of our decoding approach in integrating language models. 

Table \ref{tab_1} shows a comparison between Eesen and a hybrid HMM/DNN system. The hybrid system is constructed by following the standard Kaldi recipe ``s5'' \cite{povey2011kaldi}. Inputs of the DNN model are 11 neighboring frames of filterbank features. The DNN has 6 hidden layers and 1024 units at each layer. This DNN model contains slightly more parameters (9.2 vs 8.5 million) than the Eesen RNN model. Parameters of the DNN are initialized with restricted Boltzmann machines (RBMs) that are pre-trained in a greedy layerwise fashion \cite{hinton2006fast}. The DNN is fine-tuned to optimize the CE objective with respect to 3421 senones. For fair evaluations, we decode the DNN model using the original lexicon, rather than the expanded lexicon used by the Kaldi recipe. From Table \ref{tab_1}, we observe that the performance of the Eesen system is still behind the hybrid HMM/DNN system. Our recent developments of Eesen reveal that CTC-trained models outperform the existing hybrid systems on large-sized datasets, e.g., Switchboard. Interested readers may refer to the Eesen repository for the updates. 

A major advantage of Eesen compared with the hybrid approach is the decoding speed. The acceleration comes from the drastic reduction of the number of states, i.e., from thousands of senones to tens of phonemes/characters. To verify this, Table \ref{tab_2} compares the decoding speed of the Eesen and the hybrid HMM/DNN systems under their best decoding settings. From their real-time factors, we observe that decoding in Eesen is 3.2$\times$ faster than that of  HMM/DNN. Also, the decoding graph (TLG) in Eesen is significantly smaller than the graph (HCLG) used by HMM/DNN, which saves the disk space for storing the graphs. 
\begin{table}[th]
  \caption{\label{tab_1} {\it Performance of the phoneme-based Eesen system, and its comparison with  the hybrid HMM/DNN system built with Kaldi. ``\#Param'' means the number of parameters.  }}
  \vspace{2mm}
  \centerline{
    \begin{tabular}{| c | c | c | c | c |}
    \hline
    Model & LM & \#Param & WER\% \\
    \hline \hline
     Eesen RNN & lexicon & 8.5M & 26.92 \\
     \hline
     Eesen RNN & trigram & 8.5M & 7.87 \\
     \hline \hline
     Hybrid HMM/DNN & trigram & 9.2M & 7.14 \\
    \hline
    \end{tabular}
  }
  \vspace{-0.2in}
\end{table}

\begin{table}[th]
  \caption{\label{tab_2} {\it Comparisons of decoding speed between the phoneme-based Eesen system and the hybrid HMM/DNN system. ``RTF'' refers to the real-time factor in decoding. ``Graph Size'' means the size of the decoding graph in terms of megabytes.}}
  \vspace{2mm}
  \centerline{
    \begin{tabular}{| c | c | c | c | c |}
    \hline
    Model & RTF & Graph Size \\
    \hline \hline
     Eesen RNN & 0.64 & 263 \\
     \hline
     Hybrid HMM/DNN & 2.06 & 480 \\
    \hline
    \end{tabular}
  }
  \vspace{-0.1in}
\end{table}

\subsection{Character-based Systems} \label{sec:exp_3}

We apply the same RNN architecture discussed in Section \ref{sec:exp_2} to  modeling characters. We take the word list from the CMU dictionary as our vocabulary, ignoring the word pronunciations. CTC training deals with 59 labels including letters, digits, punctuation marks, etc. Table \ref{tab_3} shows that with the standard language model, the character-based system gets the WER of 9.07\%. 

CTC experiments in past work \cite{graves2014towards} have adopted an expanded vocabulary, and re-trained the language model using text data released together with the WSJ corpus. For fair comparison, we follow the identical configuration. OOV words that occur at least twice in the language model training texts are added to the vocabulary. A new trigram language model is built (and then pruned) with the language model training texts. Under this setup, the WER of the Eesen character-based system is reduced to 7.34\%. 

Table \ref{tab_3} lists the results of end-to-end ASR systems that have been reported in the previous work \cite{graves2014towards, hannun2014first} and on the same dataset. Our Eesen framework outperforms both \cite{graves2014towards} and \cite{hannun2014first} in terms of WERs on the testing set. It is worth pointing out that the 8.7\% WER reported in \cite{graves2014towards} is obtained not in a purely end-to-end manner. Instead, the authors of \cite{graves2014towards} generate a n-best list of hypotheses from a hybrid DNN model, and apply the CTC model to rescore the hypotheses candidates. Our Eesen numbers, in contrast, come from a completely end-to-end pipeline, without any intervention from GMM or hybrid DNN models. 
\vspace{-0.1in}
\begin{table}[th]
  \caption{\label{tab_3} {\it Performance of the character-based Eesen system using different vocabularies and language models, and its comparison with results presented in previous work. }}
  \vspace{2mm}
  \centerline{
    \begin{tabular}{| c | c | c | c |}
    \hline
    System & Vocabulary & Language Model & WER\% \\
    \hline \hline
     Eesen & Original  & Standard & 9.07 \\
     \hline
     Eesen & Expanded & Re-trained & 7.34 \\
     \hline \hline
     Graves \textit{et al.} \cite{graves2014towards} & Expanded & Re-trained & 8.7 \\
     \hline
     Hannun \textit{et al.} \cite{hannun2014first} & Original & Unknown & 14.1 \\
    \hline
    \end{tabular}
  }
  \vspace{-0.15in}
\end{table}

\section{Conclusions and Future Work} \label{sec:conclude}

In this work, we present our Eesen framework to build end-to-end ASR systems. Eesen exploits deep RNNs as the acoustic models and CTC as the training objective function. We train the RNN models in a single step, and thus are able to reduce the complexity of ASR system development. The WFST-based decoding enables efficient and effective incorporation of lexicions and language models. Because of its open-source property, Eesen can serve as a shared benchmark platform for research on end-to-end ASR.

In our future work, we plan to further improve the WERs of Eesen systems via more advanced learning techniques (e.g., expected transcription loss in \cite{graves2014towards}) and alternative decoding approach (e.g., dynamic decoders \cite{soltau2001one}). Also, we are interested to apply Eesen to various languages \cite{miao2014distributed, miao2014improving, li2015towards} and different types (e.g., noisy, far-field) of speech, and investigate how end-to-end ASR performs under these conditions. Moreover, due to the removal of GMMs, acoustic modeling in Eesen cannot leverage speaker adapted front-ends. We will study new speaker adaptation \cite{liao2013speaker, yao2012adaptation} and adaptive training \cite{miao2014towards, miao2015speaker} techniques for the CTC models.

\section{Acknowledgements}
  
This work used the Extreme Science and Engineering Discovery Environment (XSEDE), which is supported by National Science Foundation grant number OCI-1053575. This research was performed as part of the Speech Recognition Virtual Kitchen project, which is supported by the United States National Science Foundation under grant number CNS-1305365. This work was partially funded by Facebook, Inc. The views and conclusions contained herein are those of the authors and should not be interpreted as necessarily representing the official policies or endorsements, either expressed or implied, of Facebook, Inc.

%\section{REFERENCES}
\label{sec:ref}

\bibliographystyle{IEEEbib}
%\bibliography{strings,refs}
\bibliography{mybib}

\begin{thebibliography}{10}

\bibitem{dahl2012context}
George~E Dahl, Dong Yu, Li~Deng, and Alex Acero,
\newblock ``Context-dependent pre-trained deep neural networks for
  large-vocabulary speech recognition,''
\newblock {\em Audio, Speech, and Language Processing, IEEE Transactions on},
  vol. 20, no. 1, pp. 30--42, 2012.

\bibitem{hinton2012deep}
Geoffrey Hinton, Li~Deng, Dong Yu, George~E Dahl, Abdel-rahman Mohamed, Navdeep
  Jaitly, Andrew Senior, Vincent Vanhoucke, Patrick Nguyen, Tara~N Sainath,
  et~al.,
\newblock ``Deep neural networks for acoustic modeling in speech recognition:
  The shared views of four research groups,''
\newblock {\em Signal Processing Magazine, IEEE}, vol. 29, no. 6, pp. 82--97,
  2012.

\bibitem{seide2011feature}
Frank Seide, Gang Li, Xie Chen, and Dong Yu,
\newblock ``Feature engineering in context-dependent deep neural networks for
  conversational speech transcription,''
\newblock in {\em Automatic Speech Recognition and Understanding (ASRU), 2011
  IEEE Workshop on}. IEEE, 2011, pp. 24--29.

\bibitem{senior2014gmm}
Andrew Senior, Georg Heigold, Michiel Bacchiani, and Hank Liao,
\newblock ``{GMM}-free {DNN} training,''
\newblock in {\em Acoustics, Speech and Signal Processing (ICASSP), 2014 IEEE
  International Conference on}. IEEE, 2014, pp. 5639--5643.

\bibitem{bacchiani2014asynchronous}
Michiel Bacchiani, Andrew Senior, and Georg Heigold,
\newblock ``Asynchronous, online, {GMM}-free training of a context dependent
  acoustic model for speech recognition,''
\newblock in {\em Fifteenth Annual Conference of the International Speech
  Communication Association (INTERSPEECH)}. ISCA, 2014.

\bibitem{graves2014towards}
Alex Graves and Navdeep Jaitly,
\newblock ``Towards end-to-end speech recognition with recurrent neural
  networks,''
\newblock in {\em Proceedings of the 31st International Conference on Machine
  Learning (ICML-14)}, 2014, pp. 1764--1772.

\bibitem{hannun2014deepspeech}
Awni Hannun, Carl Case, Jared Casper, Bryan Catanzaro, Greg Diamos, Erich
  Elsen, Ryan Prenger, Sanjeev Satheesh, Shubho Sengupta, Adam Coates, et~al.,
\newblock ``Deepspeech: Scaling up end-to-end speech recognition,''
\newblock {\em arXiv preprint arXiv:1412.5567}, 2014.

\bibitem{hannun2014first}
Awni~Y Hannun, Andrew~L Maas, Daniel Jurafsky, and Andrew~Y Ng,
\newblock ``First-pass large vocabulary continuous speech recognition using
  bi-directional recurrent {DNNs},''
\newblock {\em arXiv preprint arXiv:1408.2873}, 2014.

\bibitem{chorowski2014end}
Jan Chorowski, Dzmitry Bahdanau, Kyunghyun Cho, and Yoshua Bengio,
\newblock ``End-to-end continuous speech recognition using attention-based
  recurrent {NN}: First results,''
\newblock {\em arXiv preprint arXiv:1412.1602}, 2014.

\bibitem{maas2015lexicon}
Andrew~L Maas, Ziang Xie, Dan Jurafsky, and Andrew~Y Ng,
\newblock ``Lexicon-free conversational speech recognition with neural
  networks,''
\newblock in {\em Proceedings of the 2015 Conference of the North American
  Chapter of the Association for Computational Linguistics: Human Language
  Technologies}, 2015.

\bibitem{bahdanau2015end}
Dzmitry Bahdanau, Jan Chorowski, Dmitriy Serdyuk, Philemon Brakel, and Yoshua
  Bengio,
\newblock ``End-to-end attention-based large vocabulary speech recognition,''
\newblock {\em arXiv preprint arXiv:1508.04395}, 2015.

\bibitem{chan2015listen}
William Chan, Navdeep Jaitly, Quoc~V Le, and Oriol Vinyals,
\newblock ``Listen, attend and spell,''
\newblock {\em arXiv preprint arXiv:1508.01211}, 2015.

\bibitem{graves2006connectionist}
Alex Graves, Santiago Fern{\'a}ndez, Faustino Gomez, and J{\"u}rgen
  Schmidhuber,
\newblock ``Connectionist temporal classification: labelling unsegmented
  sequence data with recurrent neural networks,''
\newblock in {\em Proceedings of the 23rd international conference on Machine
  learning}. ACM, 2006, pp. 369--376.

\bibitem{sak2015learning}
Hasim Sak, Andrew Senior, Kanishka Rao, Ozan Irsoy, Alex Graves, Francoise
  Beaufays, and Johan Schalkwyk,
\newblock ``Learning acoustic frame labeling for speech recognition with
  recurrent neural networks,''
\newblock in {\em Acoustics, Speech and Signal Processing (ICASSP), 2015 IEEE
  International Conference on}. IEEE, 2015, pp. 4280--4284.

\bibitem{graves2013speech}
Alex Graves, Abdel-rahman Mohamed, and Geoffrey Hinton,
\newblock ``Speech recognition with deep recurrent neural networks,''
\newblock in {\em Acoustics, Speech and Signal Processing (ICASSP), 2013 IEEE
  International Conference on}. IEEE, 2013, pp. 6645--6649.

\bibitem{graves2013hybrid}
Alex Graves, Navdeep Jaitly, and Abdel-rahman Mohamed,
\newblock ``Hybrid speech recognition with deep bidirectional {LSTM},''
\newblock in {\em Automatic Speech Recognition and Understanding (ASRU), 2013
  IEEE Workshop on}. IEEE, 2013, pp. 273--278.

\bibitem{hochreiter1997long}
Sepp Hochreiter and J{\"u}rgen Schmidhuber,
\newblock ``Long short-term memory,''
\newblock {\em Neural computation}, vol. 9, no. 8, pp. 1735--1780, 1997.

\bibitem{sak2014long}
Hasim Sak, Andrew Senior, and Fran{\c{c}}oise Beaufays,
\newblock ``Long short-term memory recurrent neural network architectures for
  large scale acoustic modeling,''
\newblock in {\em Fifteenth Annual Conference of the International Speech
  Communication Association (INTERSPEECH)}. ISCA, 2014.

\bibitem{sainath2015convolutional}
Tara~N Sainath, Oriol Vinyals, Andrew Senior, and Hasim Sak,
\newblock ``Convolutional, long short-term memory, fully connected deep neural
  networks,''
\newblock in {\em Acoustics, Speech and Signal Processing (ICASSP), 2015 IEEE
  International Conference on}. IEEE, 2015.

\bibitem{miao2015on}
Yajie Miao and Florian Metze,
\newblock ``On speaker adaptation of long short-term memory recurrent neural
  networks,''
\newblock in {\em Sixteenth Annual Conference of the International Speech
  Communication Association (INTERSPEECH)}. ISCA, 2015.

\bibitem{bengio1994learning}
Yoshua Bengio, Patrice Simard, and Paolo Frasconi,
\newblock ``Learning long-term dependencies with gradient descent is
  difficult,''
\newblock {\em Neural Networks, IEEE Transactions on}, vol. 5, no. 2, pp.
  157--166, 1994.

\bibitem{gers2003learning}
Felix~A Gers, Nicol~N Schraudolph, and J{\"u}rgen Schmidhuber,
\newblock ``Learning precise timing with {LSTM} recurrent networks,''
\newblock {\em The Journal of Machine Learning Research}, vol. 3, pp. 115--143,
  2003.

\bibitem{waibel1989phoneme}
Alex Waibel, Toshiyuki Hanazawa, Geoffrey Hinton, Kiyohiro Shikano, and Kevin~J
  Lang,
\newblock ``Phoneme recognition using time-delay neural networks,''
\newblock {\em Acoustics, Speech and Signal Processing, IEEE Transactions on},
  vol. 37, no. 3, pp. 328--339, 1989.

\bibitem{hampshire1990novel}
John~B Hampshire, Alexander~H Waibel, et~al.,
\newblock ``A novel objective function for improved phoneme recognition using
  time-delay neural networks,''
\newblock {\em Neural Networks, IEEE Transactions on}, vol. 1, no. 2, pp.
  216--228, 1990.

\bibitem{sainath2014deep}
Tara~N Sainath, Brian Kingsbury, George Saon, Hagen Soltau, Abdel-rahman
  Mohamed, George Dahl, and Bhuvana Ramabhadran,
\newblock ``Deep convolutional neural networks for large-scale speech tasks,''
\newblock {\em Neural Networks}, 2014.

\bibitem{rabiner1989tutorial}
Lawrence~R Rabiner,
\newblock ``A tutorial on hidden {Markov} models and selected applications in
  speech recognition,''
\newblock {\em Proceedings of the IEEE}, vol. 77, no. 2, pp. 257--286, 1989.

\bibitem{mohri2002weighted}
Mehryar Mohri, Fernando Pereira, and Michael Riley,
\newblock ``Weighted finite-state transducers in speech recognition,''
\newblock {\em Computer Speech \& Language}, vol. 16, no. 1, pp. 69--88, 2002.

\bibitem{povey2011kaldi}
Daniel Povey, Arnab Ghoshal, Gilles Boulianne, Luk{\'a}{\v{s}} Burget,
  Ond{\v{r}}ej Glembek, Nagendra Goel, Mirko Hannemann, Petr
  Motl{\'\i}{\v{c}}ek, Yanmin Qian, Petr Schwarz, Jan Silovsk{\'y}, Georg
  Stemmer, and Karel Vesel{\'y},
\newblock ``The {Kaldi} speech recognition toolkit,''
\newblock in {\em Automatic Speech Recognition and Understanding (ASRU), 2011
  IEEE Workshop on}. IEEE, 2011, pp. 1--4.

\bibitem{allauzen2007openfst}
Cyril Allauzen, Michael Riley, Johan Schalkwyk, Wojciech Skut, and Mehryar
  Mohri,
\newblock ``{OpenFst}: A general and efficient weighted finite-state transducer
  library,''
\newblock in {\em Implementation and Application of Automata}, pp. 11--23.
  Springer, 2007.

\bibitem{hinton2006fast}
Geoffrey Hinton, Simon Osindero, and Yee-Whye Teh,
\newblock ``A fast learning algorithm for deep belief nets,''
\newblock {\em Neural computation}, vol. 18, no. 7, pp. 1527--1554, 2006.

\bibitem{soltau2001one}
Hagen Soltau, Florian Metze, Christian F{\"u}gen, and Alex Waibel,
\newblock ``A one-pass decoder based on polymorphic linguistic context
  assignment,''
\newblock in {\em Automatic Speech Recognition and Understanding, 2001.
  ASRU'01. IEEE Workshop on}. IEEE, 2001, pp. 214--217.

\bibitem{miao2014distributed}
Yajie Miao, Hao Zhang, and Florian Metze,
\newblock ``Distributed learning of multilingual {DNN} feature extractors using
  {GPU}s,''
\newblock in {\em Fifteenth Annual Conference of the International Speech
  Communication Association (INTERSPEECH)}. ISCA, 2014.

\bibitem{miao2014improving}
Yajie Miao and Florian Metze,
\newblock ``Improving language-universal feature extraction with deep maxout
  and convolutional neural networks,''
\newblock in {\em Fifteenth Annual Conference of the International Speech
  Communication Association (INTERSPEECH)}. ISCA, 2014.

\bibitem{li2015towards}
Jie Li, Heng Zhang, Xinyuan Cai, and Bo~Xu,
\newblock ``Towards end-to-end speech recognition for chinese mandarin using
  long short-term memory recurrent neural networks,''
\newblock in {\em Sixteenth Annual Conference of the International Speech
  Communication Association (INTERSPEECH)}. ISCA, 2015.

\bibitem{liao2013speaker}
Hank Liao,
\newblock ``Speaker adaptation of context dependent deep neural networks,''
\newblock in {\em Acoustics, Speech and Signal Processing (ICASSP), 2013 IEEE
  International Conference on}. IEEE, 2013, pp. 7947--7951.

\bibitem{yao2012adaptation}
Kaisheng Yao, Dong Yu, Frank Seide, Hang Su, Li~Deng, and Yifan Gong,
\newblock ``Adaptation of context-dependent deep neural networks for automatic
  speech recognition,''
\newblock in {\em 2012 IEEE Spoken Language Technology Workshop (SLT)}. IEEE,
  2012.

\bibitem{miao2014towards}
Yajie Miao, Hao Zhang, and Florian Metze,
\newblock ``Towards speaker adaptive training of deep neural network acoustic
  models,''
\newblock in {\em Fifteenth Annual Conference of the International Speech
  Communication Association (INTERSPEECH)}. ISCA, 2014.

\bibitem{miao2015speaker}
Yajie Miao, Hao Zhang, and Florian Metze,
\newblock ``Speaker adaptive training of deep neural network acoustic models
  using i-vectors,''
\newblock {\em IEEE/ACM Transactions on Audio, Speech and Language Processing
  (TASLP)}, vol. 23, no. 11, pp. 1938--1949, 2015.

\end{thebibliography}

\end{document}